\pdfoutput=1

\documentclass[letterpaper, 10pt, conference]{ieeeconf}
\IEEEoverridecommandlockouts
\newcommand{\ours}{CoDance\xspace}

\newcommand{\oursTitle}{\ours: Learning Reactive and Compliant Human-Humanoid Interaction from Video}

\newcommand{\projecturl}{https://generalroboticslab.com/CoDance}

\usepackage[utf8]{inputenc}
\usepackage[T1]{fontenc}

\usepackage{amsmath}
\usepackage{amssymb}
\usepackage{amsfonts}
\usepackage{bm}
\usepackage{mathrsfs}
\usepackage{nicefrac}
\usepackage{cancel}

\usepackage{graphicx}
\usepackage{float}
\usepackage{subcaption}
\usepackage{wrapfig}

\usepackage[table,dvipsnames]{xcolor}

\definecolor{cvprblue}{HTML}{3576B5}

\definecolor{dukeroyal}{HTML}{00539B}

\definecolor{forest}{HTML}{3F7E31}

\definecolor{brick}{HTML}{C43119}

\makeatletter
\newcommand{\definepreset}[6]{\expandafter\def\csname preset@#1\endcsname{\colorlet{linkcolor}{#2}\colorlet{citecolor}{#3}\colorlet{urlcolor}{#4}\colorlet{deepgreen}{#5}\colorlet{deepred}{#6}}}
\newcommand{\usepreset}[1]{\@ifundefined{preset@#1}{\PackageError{palette}{No preset named '#1'}{Defined presets live in palette.tex; check the spelling.}}{\csname preset@#1\endcsname}}
\makeatother

\definepreset{cvpr}      {cvprblue}  {cvprblue}  {cvprblue}  {forest}{brick}

\usepreset{cvpr}

\usepackage{booktabs}
\usepackage{multirow}
\usepackage{tabularx}
\usepackage{makecell}
\usepackage{adjustbox}

\usepackage{algorithm}
\usepackage{algpseudocode}

\usepackage{siunitx}
\usepackage{xspace}
\usepackage{pifont}
\usepackage{balance}
\usepackage{microtype}

\let\labelindent\relax
\usepackage{enumitem}

  \usepackage{cite}

      \PassOptionsToPackage{pagebackref=true}{hyperref}

  \usepackage[breaklinks=true,colorlinks=true,
              linkcolor=linkcolor,citecolor=citecolor,urlcolor=urlcolor,
              pdfborder={0 0 0},bookmarks=true]{hyperref}

\usepackage{cleveref}

\crefformat{equation}{(#2#1#3)}
\Crefformat{equation}{Equation~(#2#1#3)}
\crefname{section}{Section}{Sections}
\Crefname{section}{Section}{Sections}
\crefname{table}{Table}{Tables}
\Crefname{table}{Table}{Tables}
\crefname{figure}{Fig.}{Figs.}
\Crefname{figure}{Fig.}{Figs.}
\crefname{algorithm}{Algorithm}{Algorithms}
\Crefname{algorithm}{Algorithm}{Algorithms}

\colorlet{ourscolor}{dukeroyal}
\newcommand{\oursline}[1]{\textcolor{ourscolor}{#1}}

\title{\LARGE \bf \oursTitle}

  \author{Zhuoqun Chen$^{1}$, Shucheng Jia$^{1}$, and Boyuan Chen$^{1}$\\
    {\small\url{\projecturl}}\thanks{$^{1}$All authors are with Duke University.}}
\begin{document}
  \bstctlcite{IEEEexample:BSTcontrolfull}

\IEEEaftertitletext{\begin{center}
  \centering
  \captionsetup{type=figure}
  \includegraphics[width=0.95\linewidth]{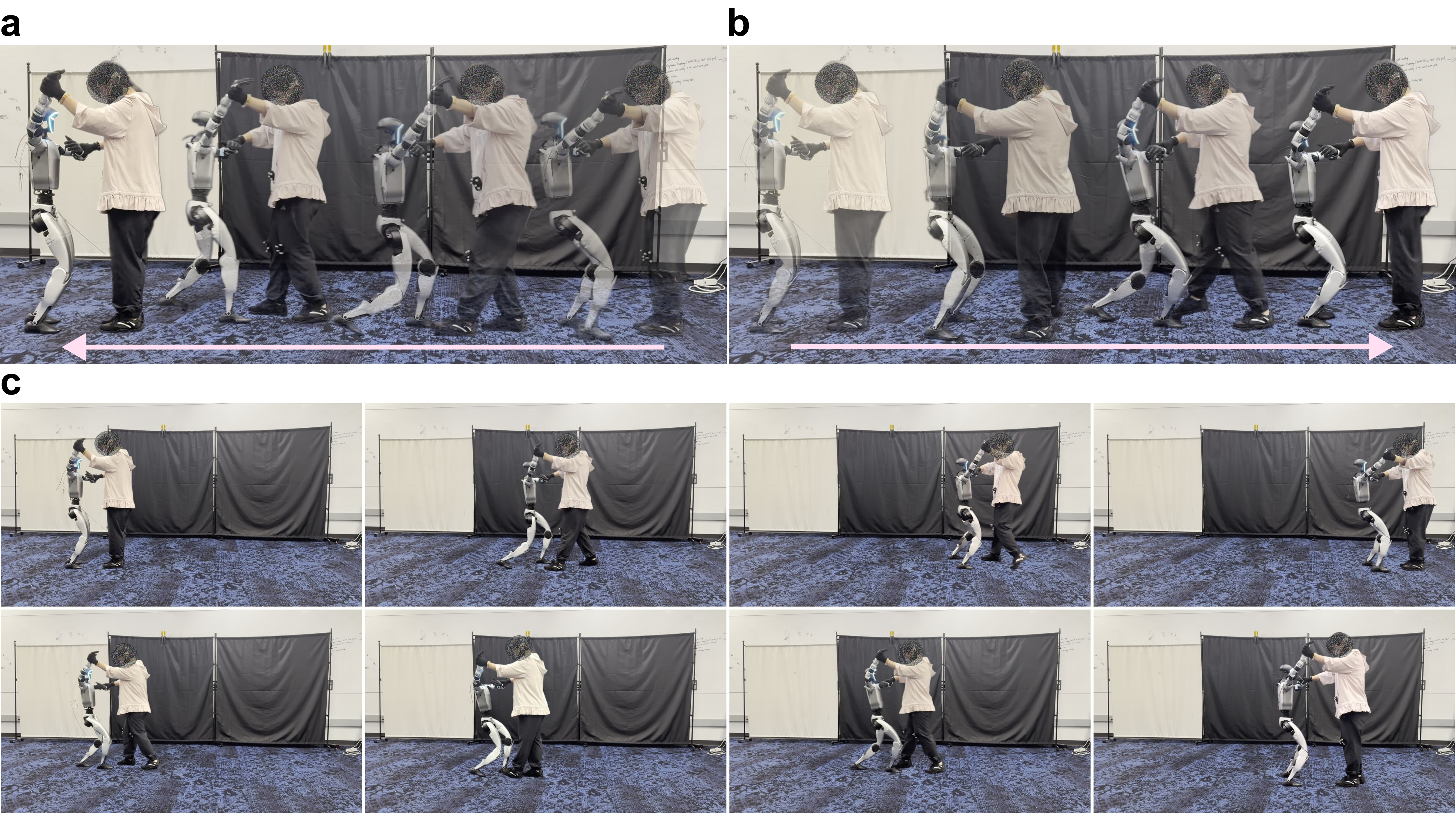}
  \caption{\textbf{Reactive and compliant human-humanoid dancing with \textsc{CoDance}.} Learned from a single video of two human dancers, the humanoid coordinates its locomotion with a human partner while maintaining compliant two-hand contact, and adapts its upper-body motion to forces transmitted through the hands. It steps backward as the partner advances (a) and forward as the partner retreats (b), and it repeats the transition between the two directions over one continuous dance (c).}
  \label{fig:teaser}
\end{center}}
\maketitle
\thispagestyle{empty}
\pagestyle{empty}

\begin{abstract}
\label{abstract}
Partnered human-humanoid interaction couples locomotion with continuous physical contact. A humanoid needs to coordinate with a person's motion while responding to interaction forces and maintaining stable and natural movement. We present \textsc{CoDance}, a framework for learning reactive and compliant human-humanoid interaction from video. We study partnered dancing as a challenging instantiation, where a humanoid coordinates its footsteps with a moving partner and maintains continuous two-hand contact. Given a single video of two human dancers, \textsc{CoDance} retargets their motions into a robot reference and a moving partner. We introduce a multi-link compliance augmentation that transforms the kinematic demonstration into force-aware training data by adapting the robot reference under structured forces at both hands. Policies trained on this data follow the observed partner while preserving the demonstrated locomotion style and responding compliantly to physical interaction. In simulation, the policies adapt their footsteps to changes in the partner and reproduce approximately 80\% of the wrist displacement encoded by the augmented demonstrations. On a physical humanoid, \textsc{CoDance} enables sustained two-hand dancing with a human partner including repeated transitions between forward and backward motions.
\end{abstract}
\section{Introduction}
\label{sec:introduction}

Humanoid robots are increasingly capable of reproducing dynamic and expressive human motion through advances in motion imitation, retargeting, and reinforcement learning~\cite{exbody,exbody2,amp}. However, most learned humanoid behaviors remain focused on the robot in isolation, tracking a motion reference or following a locomotion command while treating external contact primarily as a disturbance. Physical interaction with a person poses a unique challenge. When two partners walk, carry, guide, or dance together, their motions are continuously coupled through both observation and contact. The robot needs to adapt its motion to the partner and accommodate interaction forces while preserving balance and the natural movement.

We study this problem through partnered dancing, a demanding form of sustained physical human-humanoid interaction. The robot coordinates its footsteps with a moving partner while preserving the demonstrated locomotion style and maintaining continuous two-hand contact. Differences in timing, position, and body geometry therefore need to be accommodated through compliant motion. This creates a direct conflict with conventional motion imitation, where deviations from the reference are treated as tracking errors. Recent compliant tracking methods~\cite{softmimic} address physical perturbations during motion imitation, but partnered interaction also requires the robot to adapt its locomotion to another person while continuously responding to physical contact.

We introduce \ours, a framework for learning reactive and compliant human-humanoid interaction from a single video of two people dancing (\cref{fig:overview}). \ours reconstructs and separately retargets the two dancers, using one as the robot reference and the other as a moving partner during training. The policy observes sparse partner keypoints and coordinates its footsteps directly with the partner's motion, without relying on a prescribed velocity or trajectory. At deployment, the simulated partner is replaced by a real person whose corresponding keypoints are measured from a small set of motion-capture markers.

A central challenge is that video provides motion without the interaction forces. We address this with a multi-link compliance augmentation that transforms the kinematic demonstration into force-aware training data. Structured interaction events apply forces at one or both hands, with coupled displacements and force directions informed by the partner's motion. Each event adapts the robot reference according to the sampled force and commanded stiffness, encoding how the robot should respond to physical interaction while preserving the demonstrated motion. Temporal reversal, force-free samples, and clip chaining further expand a short demonstration into long-horizon interaction trajectories. Our final data can support different locomotion objectives while sharing the same partner-following and compliance objectives.

We evaluate each component of \ours in simulation and deploy the learned policy on a physical humanoid. Changing or removing the partner observation shows that the policy uses the observed human motion to adapt its footsteps. Under interaction forces, the learned policy reproduces approximately 80\% of the wrist displacement encoded by the augmented demonstrations, compared with about 20\% for a stiff tracking baseline. The demonstrated locomotion style can be learned through either adversarial imitation or reference tracking, while removing the locomotion objective substantially degrades the resulting motion. Clip chaining further enables transitions between forward and backward motions. We show that \ours supports a humanoid for continuous two-hand dancing with a human partner across repeated forward and backward motions. Our contributions are:
\begin{itemize}[leftmargin=*,noitemsep]
    \item We introduce \ours, a framework for learning partner-conditioned locomotion and compliant physical interaction from a single video of two human dancers.
    \item We develop a partner-informed multi-link compliance augmentation that transforms kinematic demonstrations into force-aware training data for continuous two-hand interaction.
    \item We demonstrate reactive partner following, compliant interaction, and learned locomotion style in simulation, and sustained partnered dancing on a physical humanoid.
\end{itemize}
\section{Related Work}
\label{sec:related}

\subsection{Partnered Dancing and Human-Humanoid Interaction}

Physical robot dancing has been studied for more than two decades. Ms DanceR \cite{kosuge2003msdancer} follows a human partner through force sensing on an omnidirectional base. Model-based humanoids have performed box-step dances by adapting their steps to contact forces \cite{kobayashi2022boxstep} or leading through compliant arm motion \cite{charbonneau2024reemc}. Recent learning-based methods generate dance motion on legged robots \cite{dfm}, interactive motions between two humanoids \cite{rhythm}, and hand-holding dances under velocity commands \cite{exbody}. Learned humanoids have also coordinated with people for object carrying \cite{cola,h2compact} and acquired interaction skills from human videos or paired human demonstrations~\cite{humanx,pair_dstar}. \ours studies a different setting where a humanoid learns the locomotion style from paired human video, follows an observed human partner, and maintains compliant two-hand contact.

\subsection{Whole-Body Motion Imitation for Humanoids}

Physics-based character control learns natural motion through reference tracking~\cite{deepmimic} or adversarial objectives~\cite{amp,ase}. These approaches have transferred to physical humanoids for expressive whole-body control~\cite{exbody,exbody2}, teleoperation~\cite{h2o,omnih2o}, and diverse motion tracking~\cite{beyondmimic}. These methods primarily consider single-agent motion, where tracking objectives keep the robot close to its reference. Physical interaction introduces contact forces that can require motion away from this reference. In \ours, the locomotion objective preserves the demonstrated lower-body motion while the upper body adapts to interaction with the partner.

\subsection{Compliant Control and Motion Tracking}

Impedance and admittance control regulate robot motion under physical contact~\cite{hogan1985impedance}. Learned controllers extend these ideas to legged robots through impedance-reference tracking~\cite{facet}, joint position-force control~\cite{unifp}, and compliant task-space control for humanoid loco-manipulation~\cite{ceer}. For humanoid motion tracking, SoftMimic~\cite{softmimic} augments references with inverse-kinematics solutions under sampled forces, while GentleHumanoid~\cite{gentlehumanoid} learns upper-body compliance using guiding forces and human motion. Our augmentation takes one step further to target sustained partnered locomotion. Force events act on one or both wrists with coupled displacements and directions informed by the partner's motion, producing force-aware references for compliant two-hand interaction during partner following.
\section{Method}
\label{sec:method}

\begin{figure*}[t]
  \centering
  \includegraphics[width=0.85\textwidth]{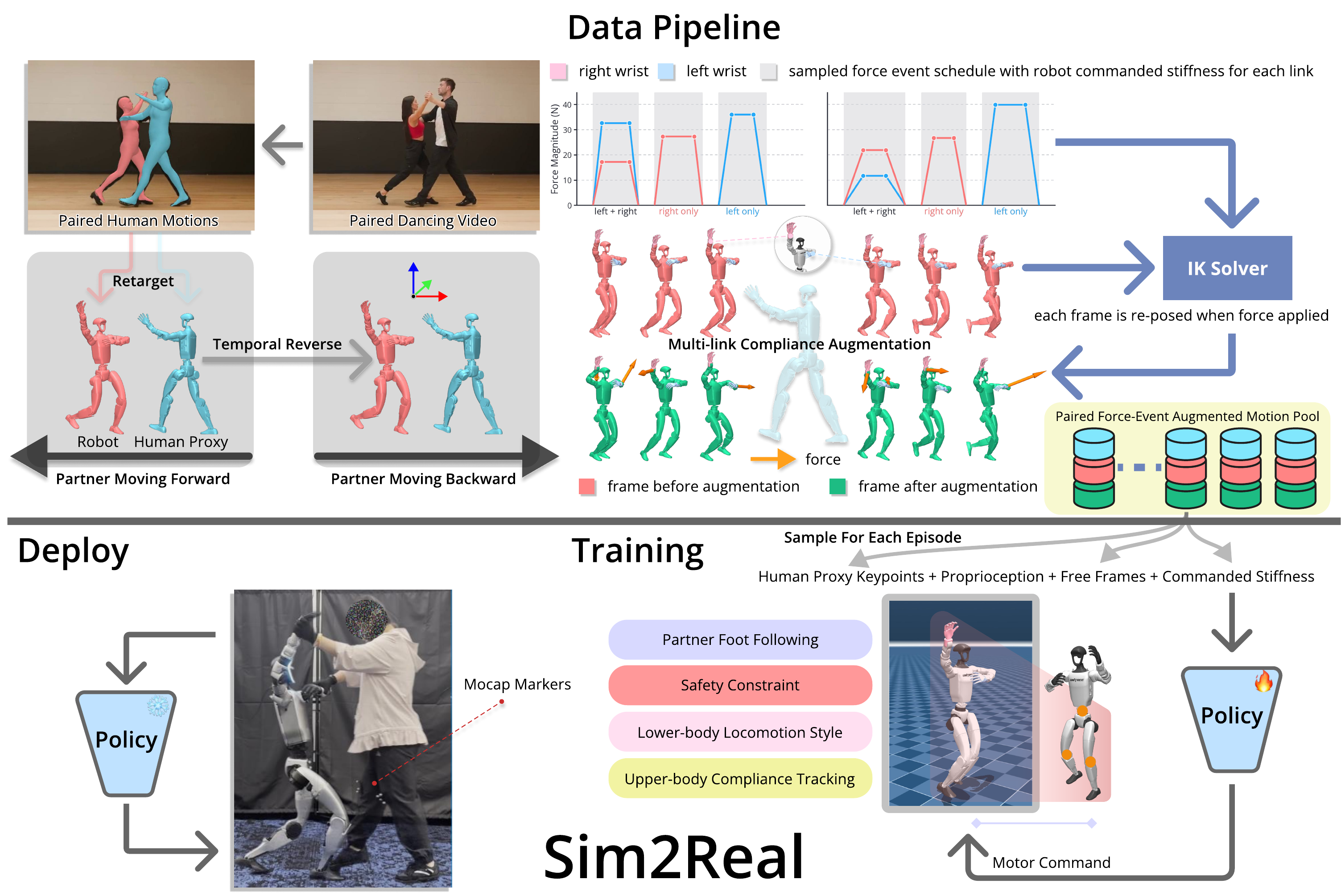}
  \caption{Overview of \textsc{CoDance}. A paired-dancing video is converted into force-aware training data through motion retargeting and multi-link compliance augmentation. The policy learns to follow the observed partner, maintain the demonstrated locomotion style, and remain compliant at both hands. At deployment, the simulated partner is replaced by a person. Training is illustrated with the decoupled policy.}
  \label{fig:overview}
\end{figure*}

\ours consists of three stages (\cref{fig:overview}). The data pipeline turns a paired-dancing video into force-aware training data (\cref{sec:data}). A policy then learns partner following and two-hand compliance together with a locomotion objective  (\cref{sec:policies}). At deployment, a person replaces the simulated partner and the policy runs directly on a physical humanoid (\cref{sec:deploy}).

\subsection{Data Pipeline}
\label{sec:data}

\noindent\textbf{From Video to a Paired Clip.} PromptHMR \cite{prompthmr} recovers the SMPL-X \cite{smplx} motions of both dancers, and GMR \cite{retargeting_matters} retargets each motion separately to the humanoid. One dancer provides the robot reference and the other provides the partner motion. Since separate retargeting does not preserve their relative placement, we position the partner using a fixed following offset. During training, a second humanoid kinematically replays this motion as a human proxy. We also reverse the \SI{9.7}{\second} paired clip in time to cover both directions of travel. The video does not carry contact forces, so the proxy's upper body does not collide with the robot, and the hand forces come from the force augmentation below. Its lower body retains collision so that stepping into the partner can be penalized.

\noindent\textbf{Multi-Link Compliance Augmentation.} Our augmentation builds on the quasi-static compliance solve of SoftMimic \cite{softmimic}. Force events are distributed throughout the clip, with each event following a ramp, hold, and release profile (\cref{fig:overview}). For a force $f$ and robot stiffness $K$, the contacted link is displaced by $f/K$. We sample $K$ log-uniformly from \SIrange{10}{1000}{\newton\per\meter}. A whole-body inverse-kinematics then adapts the reference while holding the feet fixed, constraining the center of mass, and weakly tracking the remaining body links. If a frame violates feasibility or self-collision constraints, the event is recomputed with the wrist force directed away from the torso or reduced in magnitude. We call the resulting motion adapted reference and the original motion the free reference. We extend the sampled interaction beyond the single-link perturbations used in SoftMimic in three ways.

\textbf{Coupled two-hand interaction.} Instead of SoftMimic's left or right wrist, each event samples an interaction intent that sets the contacted wrists and their relative displacements. The intents include a single wrist (\texttt{single}), common motion of both wrists (\texttt{along}), mirrored motion (\texttt{open}), opposing vertical motion (\texttt{twist}), and motion toward or away from each other (\texttt{oppose}). Each contacted wrist has an independently sampled stiffness, except under \texttt{oppose}, where both wrists share the same stiffness to produce equal and opposite sampled forces.

\textbf{Partner-guided force direction.}
The partner's smoothed horizontal velocity sets the base direction of each event. Most events push along this velocity, and the rest push along the robot's facing axis against the direction of travel. When the partner is nearly still, the direction is drawn at random. Directional jitter expands each base direction into a local cone.

\textbf{Balance-aware force scaling.}
A force applied at the hands requires a corresponding center-of-mass shift for balance. Before solving the event, we scale its force once so that the required shift stays within a prescribed budget throughout the event. This preserves the sampled ramp, hold, and release profile while keeping the adapted motion feasible.

\begin{algorithm}[t]
\caption{Multi-link compliance augmentation. Black: the SoftMimic
\cite{softmimic} single-link procedure as we implement it; \oursline{blue}: ours.}
\label{alg:compliance}
\begin{algorithmic}[1]
\Require reference clip; contactable links (the two wrists); \oursline{partner clip}; stiffness range; force ceiling; \oursline{balance budget}; displacement scale
\Ensure  adapted clip; force schedule
\State $t \gets 0$; events $\gets \varnothing$
\While{$t <$ clip length}
  \State $t \gets t + \mathcal{U}(0.5, 1.5)$\,s; skip if a wrist is busy
  \State \oursline{draw the ramp and hold durations}
  \State \oursline{draw an intent; it fixes which links are contacted (one or both wrists) and how their displacements relate}
  \State \oursline{draw the force direction from the partner's motion; random if the partner is still}
  \For{ \oursline{each contacted link} }
    \State draw $K$ log-uniformly (shared by the pair under oppose) and $\delta$ from the intent and the displacement scale
    \State shrink $\delta$ until $K\|\delta\| \le$ ceiling;\quad $f \gets K\delta$
  \EndFor
  \State \oursline{scale the event once so its CoM shift stays within the budget}
  \State record the event
\EndWhile
\For{each frame}
  \State contacted links: target $=$ reference $+\, f/K$
  \State solve whole-body IK: feet held, CoM task, every other body weakly held at its reference pose
  \If{the frame fails a feasibility \oursline{or self-collision} check}
    \State \oursline{turn the wrist's force outward or} scale the event down, then re-solve from the event's start
  \EndIf
\EndFor
\State \Return adapted clip, force schedule
\end{algorithmic}
\end{algorithm}
\begin{table}[t]
\centering
\caption{Augmentation parameters.}
\label{tab:params}
\footnotesize
\begin{tabular}{@{}>{\raggedright\arraybackslash}p{0.52\linewidth}>{\raggedright\arraybackslash}p{0.42\linewidth}@{}}
\toprule
Parameter & Value \\
\midrule
Robot stiffness $K$ & log-uniform, \numrange{10}{1000} \si{\newton\per\meter} \\
Environment stiffness & same range, independent draw \\
Force ceiling & \SI{140}{\newton} \\
Displacement scale $d_{\max}$ & \SI{0.15}{\meter} \\
\quad per-event draw & $\mathcal{U}(0.03, d_{\max})$\,m \\
Balance budget (CoM shift) & \SI{0.10}{\meter} \\
Intent probability, single and oppose & $0.25$ and $0.075$ \\
\quad along, open, twist (each) & $0.225$ \\
Vertical direction scale & $0.5$ \\
Push probability $p_{\mathrm{push}}$ & $0.85$ \\
Partner velocity smoothing window & \SI{0.4}{\second} \\
Partner speed lower bound & \SI{0.1}{\meter\per\second} \\
Direction jitter & $0.4$ \\
Facing-axis displacement cap & \SI{0.08}{\meter} \\
CoM task weights (XY, Z) & $0.1$, $10^{-6}$ \\
Loading rate limit & \SI{1.0}{\meter\per\second} \\
Ramp, hold (\si{\second}) & $\mathcal{U}(0.2,0.6)$, $\mathcal{U}(0.4,0.8)$ \\
Clips per direction & 100 with force events, 100 force-free \\
\bottomrule
\end{tabular}
\end{table}

Each augmented clip stores its adapted reference and a force schedule containing the force, commanded stiffness and timing of every event. Force-free clips use the free reference with a commanded stiffness resampled every 2 to \SI{5}{\second}. Our dataset contains 200 clips with force events and 200 force-free clips, evenly divided between forward and reversed motion. \Cref{alg:compliance,tab:params} provide the full augmentation procedure and parameters.

\subsection{Policy Training}
\label{sec:policies}

\ours is not tied to one specific policy design. Humanoid motion imitation takes its motion style either from an adversarial reward \cite{amp} or from reference tracking~\cite{deepmimic,beyondmimic}, so we train one policy of each kind on the same data. Both policies share the observations, the environment spring, every reward term except the locomotion objective, and the training setup. We call the policy with the adversarial locomotion objective the decoupled policy, since only its upper body tracks the reference, and the policy with lower-body reference tracking the whole-body tracker.

\noindent\textbf{Observations.} The actor observes proprioception with a short history, namely the base angular velocity, the joint positions and velocities, the previous actions and the projected gravity. It also observes the free-reference anchor orientation and the joint states, which provide the phase of the source motion. Partner observations contain the pelvis and both knees in the robot pelvis frame over the last second. It also observes the commanded stiffness of each wrist. The actor does not observe the adapted reference or applied force. Its response to physical interaction therefore depends on the commanded stiffness and the resulting proprioceptive motion. The critic additionally observes base linear velocity, the adapted reference, applied and scheduled forces, environment stiffness, and the robot body poses. The policy does not take a velocity command. It follows the partner from the partner's keypoints.

\noindent\textbf{Interaction Force.} During a force event, the force on each contacted wrist is modeled as a spring toward a setpoint $x^{*}_t$,
\begin{equation}
  f_t = K_{\mathrm{env}} \bigl( x^{*}_t - x_t \bigr),
  \label{eq:spring}
\end{equation}
where $x_t$ is the wrist position and $K_{\mathrm{env}}$ is drawn log-uniformly for each contacted wrist, independently of the robot stiffness. The setpoint $x^{*}_t$ is placed so that the spring delivers the scheduled force at the adapted reference. Deviations from the adapted reference will then change the interaction force. An analogous rotational spring acts on the wrist orientation with zero scheduled torque.

\noindent\textbf{Rewards.} The reward contains partner following, compliance, locomotion, and regularization terms,
\begin{equation}
  r_t = r^{\mathrm{partner}}_t + r^{\mathrm{comp}}_t + r^{\mathrm{style}}_t + r^{\mathrm{reg}}_t ,
  \label{eq:reward}
\end{equation}
where $r^{\mathrm{style}}$ is the locomotion objective. Tracking terms use exponential kernels $\exp(-e^2/\sigma^2)$~\cite{deepmimic,beyondmimic}.

\textbf{Partner following.}
The robot follows the partner through corresponding foot targets. Each robot foot is rewarded for staying near the opposite-side partner foot with a fixed following offset,
\begin{equation}
  r^{\mathrm{foot}}_t = \sum_{s \in \{\mathrm{L},\mathrm{R}\}}
    \exp\!\left( -\frac{\bigl\| x^{s}_t - \bigl( c^{\bar{s}}_t - o \bigr) \bigr\|^2}{\sigma_{\mathrm{f}}^2} \right),
  \label{eq:foot}
\end{equation}
where $x^{s}_t$ is robot foot $s$, $c^{\bar{s}}_t$ is the opposite-side partner foot, and $o$ is a \SI{0.5}{\meter} following offset. A feet air-time term rewards natural step duration. Contact with the partner's lower body is penalized.

\textbf{Compliance.}
The upper body tracks the adapted reference under the scheduled interaction forces. For link positions,
\begin{equation}
  r^{\mathrm{arm}}_t = \exp\!\left( -\frac{1}{|\mathcal{B}_{\mathrm{up}}|} \sum_{b \in \mathcal{B}_{\mathrm{up}}}
    \frac{\bigl\| p^{b}_t - \tilde{p}^{b}_t \bigr\|^2}{\sigma_{\mathrm{u}}^2} \right),
  \label{eq:upper}
\end{equation}
where $\mathcal{B}_{\mathrm{up}}$ contains the shoulder, elbow, and wrist links. The adapted reference $\tilde{p}^{b}_t$ is re-anchored to the ground-plane position and heading of the robot torso, making the tracking error invariant to global locomotion. Link orientations are treated similarly. Additional terms place higher tracking weight on the wrists and encourage the applied wrist forces and torques to match their scheduled values.

\textbf{Locomotion objective.}
The decoupled policy uses an adversarial motion reward \cite{amp}. Its discriminator compares two-frame transitions from policy rollouts with transitions from the training references using lower-body and torso poses in the pelvis frame. Arm states are excluded so that physical interaction does not enter the locomotion-style objective. Expert transitions use adapted references for clips with force events and free references for force-free clips.

The whole-body tracker uses lower-body reference tracking in place of the adversarial reward, following the motion-tracking formulation of prior work~\cite{softmimic}. It tracks lower-body poses relative to the torso, their global velocities, and the global torso pose. Both policies use force-aware reference state initialization \cite{deepmimic,softmimic}, including initialization within active force events.

\textbf{Regularization.}
$r^{\mathrm{reg}}$ penalizes action rate, joint-limit violations, self-collisions, and the joint accelerations of the hips and knees.

\noindent\textbf{Clip-Chaining and Training Details.} A dance moves forward and backward in turn, so dancing longer than one clip requires the transition from a clip in one direction to a clip in the other. During training, clips are chained at their ends. When a clip runs out, a clip drawn at random from the training data starts from the partner's current position, so an episode spans more than one clip. Episodes last \SI{20}{\second}, about two clips. We use the domain randomization of the BeyondMimic in mjlab \cite{beyondmimic,mjlab}, including torso center-of-mass offsets, foot friction, and joint encoder bias. Random pushes are disabled because the scheduled interaction events already perturb the robot. Both policies are trained with PPO in MuJoCo through mjlab \cite{mjlab} using 16384 parallel environments at \SI{50}{\hertz} control rate. Actor and critic networks have three hidden layers with 512, 256 and 128 units. Training converges in approximately \SI{37}{\hour} for the decoupled policy and \SI{29}{\hour} for the whole-body tracker on one NVIDIA L40S.

\subsection{Hardware Deployment}
\label{sec:deploy}

We deployed the converged decoupled policy on a Unitree G1 at \SI{50}{\hertz}. Joint position targets were sent through the Unitree SDK at \SI{200}{\hertz}. A Vicon system tracked the person's pelvis and knees and the robot pelvis, providing the same partner keypoints used during training. Marker heights were calibrated once to match the simulated proxy. Both wrists used a commanded stiffness of \SI{140}{\newton\per\meter}. Physical forces came directly from the person's hands during the two-hand hold.
\section{Experiments}
\label{sec:experiments}

We evaluated four aspects of \ours in simulation: reactive partner following, compliant interaction, locomotion style, and transitions across chained clips. Targeted ablations isolated the components responsible for each behavior. We then evaluated partnered dancing with a human on the physical robot.

\subsection{Experimental Setup}

All policies were trained on the same half of the augmented clips (seen). The remaining half of the clips (unseen) were generated independently by the same augmentation pipeline with newly sampled event timing, interaction intents, stiffness commands, and forces.

We evaluated the \textbf{Decoupled} and \textbf{Whole-body} policies from \cref{sec:policies} and three baselines. \textbf{Whole-body w/o foot constraint} removed the foot-following reward in \cref{eq:foot} and the foot-distance termination.
\textbf{Stiff tracker} tracked the free reference under the same force events, without force matching or the tighter wrist-tracking terms.
\textbf{Decoupled w/o adversarial term} removed the adversarial locomotion objective.

Simulation results were averaged over three evaluation seeds with 400 trials per seed. A trial terminates if the torso height drops below \SI{0.3}{\meter}, its orientation deviates from the reference by more than \SI{1.2}{\radian}, or either foot moves more than \SI{1.2}{\meter} from its partner-relative target. Hardware evaluation follows \cref{sec:deploy}.

\subsection{Evaluation Metrics}

\begin{itemize}[leftmargin=*,noitemsep]

  \item \textbf{Partner following.} Foot error is the ground-plane distance of each foot from its follow target averaged over both feet and over time before the first termination, or read at \SI{1}{\second} after the trial starts in \cref{tab:reaction}. Root distance is the ground-plane distance between the two pelvises, with a target of \SI{0.5}{\meter} (\cref{tab:benchmark_ours}). Success rate is the fraction of trials that reach the end of the clip, or of a chain of clips, without a termination (\cref{tab:reaction,tab:chain}).

  \item \textbf{Compliance.} Wrist displacement and force error were read over the last \SI{0.2}{\second} of each force event's hold, in the torso-anchored frame. The wrist displacement measures how far the wrist moves from the free reference along the applied force, for the policy and for the adapted reference. Mean displacement error is the absolute difference between the two wrist displacements, averaged over events (\cref{fig:compliance}). Relative adapted ratio is the policy's wrist displacement summed over events, divided by the same sum for the adapted reference (\cref{fig:compliance}). Force error is the magnitude of the difference between the applied and the scheduled force, averaged over events (\cref{tab:benchmark_ours}). Upper-body tracking error is the torso-relative mean per-keybody position and rotation error of the upper body against the adapted reference (R-MPKPE, R-MPKRE), with the reference re-anchored to the robot's torso at each step as in \Cref{eq:upper} (\Cref{tab:benchmark_ours}).

  \item \textbf{Locomotion style.} Lower-body tracking error is the R-MPKPE and R-MPKRE of the lower body against the adapted reference (\Cref{tab:benchmark_ours}).

  \item \textbf{Long horizon.} We report over a chain of two clips (\Cref{sec:chain}), the success rate and the foot error before the first termination (\Cref{tab:chain}) .
\end{itemize}

\subsection{Reactive Partner Following}
\label{sec:following}

We first tested whether the robot responded to the observed partner instead of simply replaying its reference motion. We evaluated two changes to the partner observation. In the masked condition, all partner keypoints are set to zero. In the reversed condition, the partner motion is temporally reversed while the robot reference remains unchanged. We also evaluated the whole-body policy trained without the foot-following constraint.

\begin{table}[t]
  \caption{Evaluation results on different partner observation and foot constraint settings. Masking out the partner observation or reversing the partner leads to a large foot error and a large drop in success rate. Training without the foot constraint leads to a large foot error.}
  \label{tab:reaction}
  \centering
  \footnotesize
  \setlength{\tabcolsep}{3pt}
  \adjustbox{max width=\columnwidth}{\begin{tabular}{llrr}
    \toprule
    Policy & Partner & \makecell{Foot Err\\at \SI{1}{\second} (m)} & Success Rate \\
    \midrule
    Decoupled & observed & 0.057 & 99.6\,\% \\
     & masked out & 0.421 & 0.0\,\% \\
     & reversed & 0.267 & 28.6\,\% \\
    \midrule
    Whole-body & observed & 0.072 & 99.7\,\% \\
     & masked out & 0.430 & 0.0\,\% \\
     & reversed & 0.255 & 24.3\,\% \\
    \midrule
    Whole-body w/o foot constraint & observed & 0.179 & 99.6\,\% \\
    \bottomrule
  \end{tabular}}
\end{table}

As shown in \cref{tab:reaction}, both policies completed nearly every trial with the original partner observation. Masking the partner observation caused all trials to fail, with foot error reaching approximately \SI{0.42}{\meter} after \SI{1}{\second}. Since zero-valued observations were outside the training distribution, we additionally used the reversed condition to separate partner dependence from this distribution shift. Reversing the partner avoided this problem, because the policies observed the partner moving in both directions in training and only its mismatch with the robot's reference was new. With this in-distribution but mismatched input, the foot error was about \SI{0.26}{\meter} after one second and about a quarter of the trials finished the clip. Without the foot constraint, the whole-body tracker finished nearly every trial but did not follow the partner, with 2.5 times the foot error of the constrained whole-body tracker after \SI{1}{\second}. These results show that partner-conditioned locomotion depends on both the observed partner motion and the foot-following objective. The supplementary video further shows hardware behavior when the human partner pauses during the dance.

\subsection{Compliance Interaction}
\label{sec:compliance}

\begin{figure}[t]
  \centering
  \includegraphics[width=0.8\linewidth]{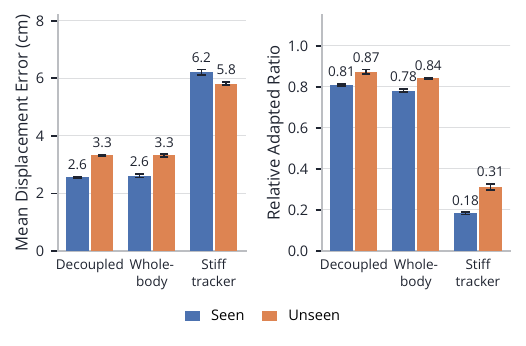}
  \caption{Evaluation results on compliance interaction. Under force, both compliant policies move the wrist most of the distance encoded in the adapted reference, while the stiff tracker moves it far less. Error bars show the standard deviation over evaluation seeds.}
  \label{fig:compliance}
\end{figure}

We next evaluated whether the force-aware training data produced compliant wrist motion during partner-conditioned locomotion, on both seen and unseen clips. The baseline was the stiff tracker, which we compared with the adapted reference of the same events.

As shown in \Cref{fig:compliance}, the two compliant policies had similar displacement error, and their relative adapted ratios differed by 0.03. On seen clips, their wrists moved about \SI{80}{\percent} as far as in the adapted reference, with a mean displacement error of \SI{2.6}{\centi\meter}. The stiff tracker moved its wrist about \SI{20}{\percent} as far, with more than twice the displacement error. On unseen clips, the relative adapted ratio was higher for every policy, including the stiff tracker. The displacement error of the two compliant policies grew by less than \SI{1}{\centi\meter}. The stiff tracker remains the farthest from the adapted reference on both measures. These results show that training with the adapted references produced the compliant wrist motion encoded by the augmentation and transferred to unseen force-event schedules.

\subsection{Locomotion Style}
\label{sec:style}

\begin{figure}[t]
  \centering
  \includegraphics[width=0.85\linewidth]{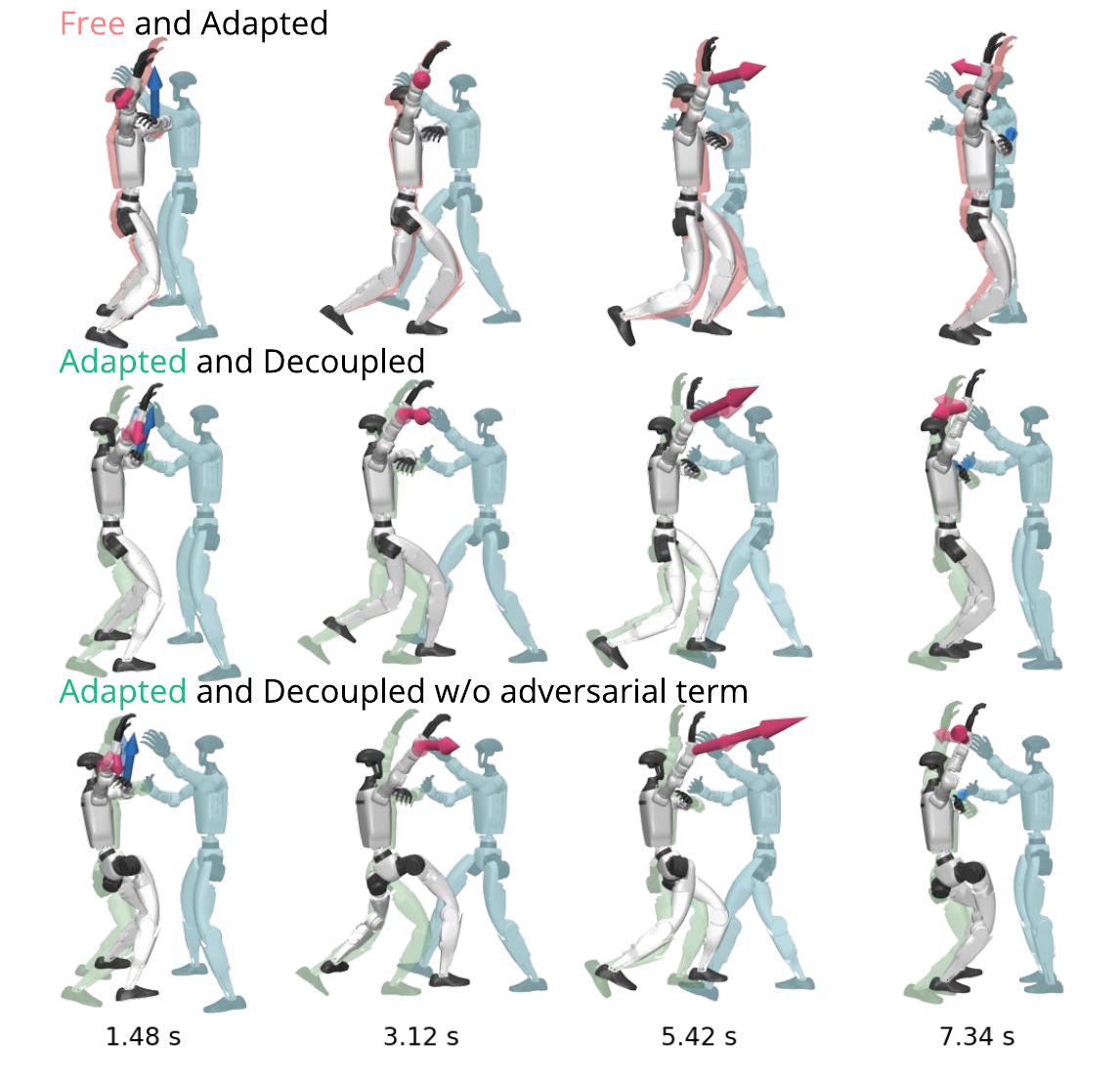}
  \caption{Visualization of the free and adapted references (top) and of the decoupled policy with and without the adversarial term, at the middle of four force holds. The partner is in teal, and arrows mark the force at the left wrist (blue) and right wrist (pink), with the applied force solid and the scheduled force faded. Without the adversarial term, the pelvis turns away from the dancer's heading, the waist and hips twist back toward the direction of travel, and the stance widens into a crouch. Top row also shows that separate retargeting does not preserve relative position in the paired motion clip.}
  \label{fig:amp_ablation}
\end{figure}

\begin{table*}[t]
  \centering
  \caption{Evaluation results on different locomotion objective settings. Removing the adversarial term from the decoupled policy leads to large lower-body tracking errors at comparable foot error. The whole-body tracker leads to tracking errors and foot error similar to the decoupled policy. The standard deviation is below 0.003 in the tracking and following columns and 0.1\,N in the force error. Bold marks the best per column, and for Root Distance the value closest to 0.5\,m.}
  \label{tab:benchmark_ours}
  \footnotesize
  \setlength{\tabcolsep}{3pt}
  \adjustbox{max width=\textwidth}{\begin{tabular}{lrrrrrrrrrrrrr}
\toprule
 & \multicolumn{4}{c}{Locomotion Style} & \multicolumn{5}{c}{Compliance} & \multicolumn{4}{c}{Partner Following} \\
\cmidrule(lr){2-5} \cmidrule(lr){6-10} \cmidrule(lr){11-14}
Methods & \multicolumn{2}{c}{\makecell{Lower\\R-MPKPE (m) $\downarrow$}} & \multicolumn{2}{c}{\makecell{Lower\\R-MPKRE (rad) $\downarrow$}} & \multicolumn{2}{c}{\makecell{Upper\\R-MPKPE (m) $\downarrow$}} & \multicolumn{2}{c}{\makecell{Upper\\R-MPKRE (rad) $\downarrow$}} & \makecell{Force Err\\(N) $\downarrow$} & \multicolumn{2}{c}{Foot Err (m) $\downarrow$} & \multicolumn{2}{c}{Root Distance (m)} \\
\cmidrule(lr){2-3} \cmidrule(lr){4-5} \cmidrule(lr){6-7} \cmidrule(lr){8-9} \cmidrule(lr){11-12} \cmidrule(lr){13-14}
 & w/ force & force-free & w/ force & force-free & w/ force & force-free & w/ force & force-free & w/ force & w/ force & force-free & w/ force & force-free \\
\midrule
Decoupled & 0.089 & 0.082 & 0.225 & 0.198 & 0.041 & 0.034 & 0.144 & 0.084 & 7.8 & \textbf{0.072} & 0.053 & \textbf{0.490} & 0.470 \\
Decoupled w/o adversarial term & 0.143 & 0.142 & 0.587 & 0.590 & \textbf{0.034} & \textbf{0.024} & 0.142 & 0.083 & 8.8 & \textbf{0.072} & \textbf{0.047} & 0.533 & \textbf{0.517} \\
Whole-body & \textbf{0.069} & \textbf{0.065} & \textbf{0.197} & \textbf{0.177} & \textbf{0.034} & 0.025 & \textbf{0.137} & \textbf{0.079} & \textbf{7.7} & 0.074 & 0.059 & 0.431 & 0.419 \\
\bottomrule
\end{tabular}
  }
\end{table*}

We tested whether the demonstrated locomotion style depended on the locomotion
objective by comparing the decoupled policy with the whole-body tracker on the unseen clips and by removing the adversarial reward from the decoupled policy, and whether changing it affects following or compliance (\cref{tab:benchmark_ours}).

The rows in \cref{tab:benchmark_ours} differ most in the locomotion style columns. The whole-body tracker tracked the lower body most closely, as its objective required, and the decoupled policy was within \SI{2}{\centi\meter} and \SI{0.03}{\radian} of it while keeping the root distance closer to the following distance. Without the adversarial reward, the decoupled policy still followed the partner with comparable foot error, but its lower-body rotation error was 2.6 times as large on clips with force events and 3.0 times as large on force-free clips. \Cref{fig:amp_ablation} visualized one seen clip with force events. Across the three rows in \Cref{tab:benchmark_ours}, the foot error, force error and upper-body tracking error stayed within about \SI{1}{\centi\meter}, \SI{0.01}{\radian} and \SI{1}{\newton} of each other, while the root distance changed by up to \SI{10}{\centi\meter}. Overall, the locomotion style came from the locomotion objective, while foot error and compliance change little when the objective was swapped or removed.

\subsection{Benefit of Clip-Chaining}
\label{sec:chain}

Clip-chaining trained the policy to continue from the end of one clip into another. We tested whether following holds across the transition between directions, on chains of two unseen clips with force events, one in each direction, played back to back in both orders. We evaluated the decoupled policy trained with and without clip-chaining, both with clip-chaining at evaluation, and the whole-body tracker also trained with clip-chaining (\cref{tab:chain}).

\begin{table}[t]
  \caption{Evaluation results on different clip-chaining settings.
  Training without clip-chaining leads to a large drop in success rate at comparable foot error.}
  \label{tab:chain}
  \centering
  \footnotesize
  \setlength{\tabcolsep}{3pt}
  \adjustbox{max width=\columnwidth}{\begin{tabular}{lrr}
    \toprule
    Policy & Foot Err (m) & Success Rate \\
    \midrule
    Decoupled trained w/ chaining & 0.073 & 97.6\,\% \\
    Decoupled trained w/o chaining & 0.076 & 48.0\,\% \\
    Whole-body trained w/ chaining & 0.074 & 96.2\,\% \\
    \bottomrule
  \end{tabular}}
\end{table}

Both policies trained with clip-chaining succeeded in nearly every trial with similar foot error as on a single clip (\cref{tab:benchmark_ours}). The policy trained without clip-chaining succeeded in about half of the trials. Its foot error before termination matched the other two, so it did not lose the partner before failing. It failed at the clip switch while still following. Overall, following held over an unseen chain for both locomotion objectives when training with clip-chaining and this mechanism maintained partner following across changes in travel direction.

\subsection{Hardware Results}
\label{sec:hardware}

\begin{figure}[t]
  \centering
  \includegraphics[width=0.85\linewidth]{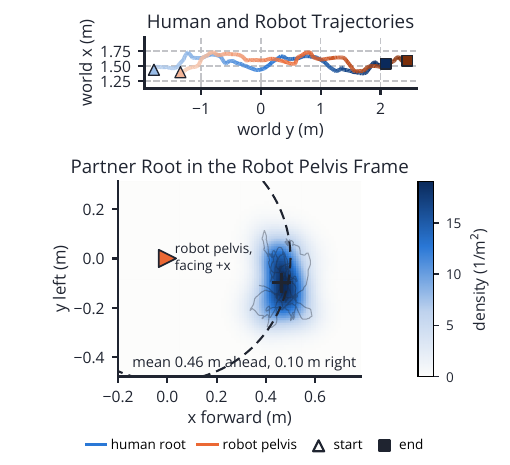}
  \caption{Visualization of one real-robot dance with a human partner. The robot keeps close to the following distance for most of the dance. Top: the trajectories of the robot and the partner. Bottom: the partner's root in the robot's pelvis frame.}
  \label{fig:real_run}
\end{figure}

\begin{figure}[t]
  \centering
  \includegraphics[width=\linewidth]{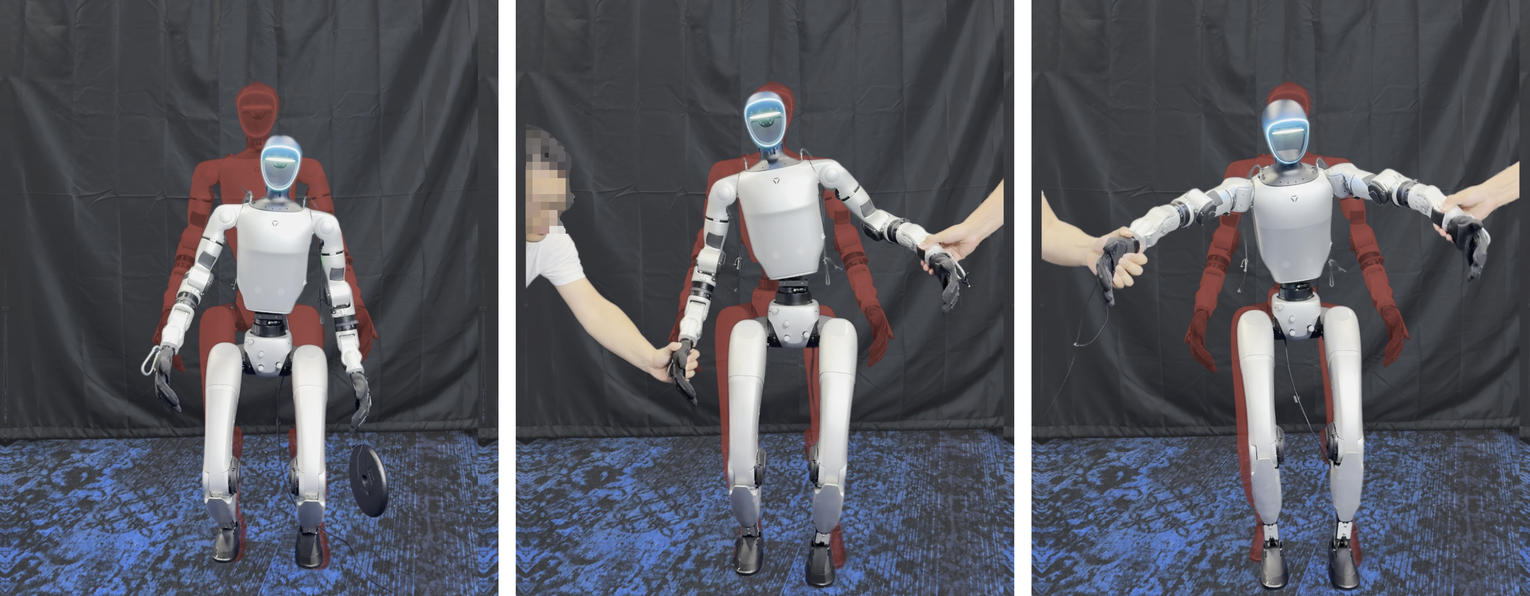}
  \caption{Visualization of a standing policy under force on one or both hands.
    The red overlay is the reference standing pose. Left: one hand loaded by a hanging weight. Middle: the two hands pulled in different directions. Right: both hands pulled outward.
    The arms move with the pull and the feet stay still.}
  \label{fig:stand_drag}
\end{figure}

We evaluated the Decoupled policy (\cref{fig:real_run} and the supplementary video) with a human partner using the setup in \cref{sec:deploy}. The robot maintains two-hand contact and follows the partner without falling in all three repeated runs. One forward run lasts \SI{12.8}{\second}, while two \SI{38.9}{\second} runs alternate forward and reversed clips until the reference ends. The root distance remains within \SI{0.1}{\meter} of the \SI{0.5}{\meter} target for 93\%, 83\%, and 73\% of the three runs, with mean distances between \SI{0.48}{\meter} and \SI{0.50}{\meter}. The pair travels approximately \SI{3.8}{\meter} during the run in \cref{fig:real_run}.
We also qualitatively evaluated a standing policy trained with the same augmentation pipeline without the partner guidance as shown in \Cref{fig:stand_drag}. It illustrated the compliant response at the hands.
The supplementary video shows the complete hardware interactions.
\section{Conclusion, limitations, and future work}
\label{sec:conclusion}

We presented \textsc{CoDance}, a framework for learning reactive and compliant human-humanoid dancing from a single video of two human dancers. \textsc{CoDance} converts kinematic demonstrations into force-aware training data and learns a policy that coordinates its footsteps with an observed partner while maintaining compliant two-hand interaction and the demonstrated locomotion style. In simulation, the learned policies reproduce most of the wrist displacement encoded by the augmented references and maintain partner following across chained clips. On a physical humanoid, \textsc{CoDance} enables sustained two-hand dancing with a human partner across forward and backward motions.

Our study is currently limited to one video and one dance. The simulated partner replays a fixed motion, with interaction forces generated by the augmentation. The policies still depend on their reference clip, so only about a quarter of the trials finish a clip when the partner moves opposite to it. The foot-following target also keeps the robot at a fixed offset along a single direction from the partner, so dances that change the partners' relative placement are not covered. In future work, we will study rise and fall, turning and the box step, and extend the augmentation to any selected link.
\space
\section*{Acknowledgment}

This work is supported by DARPA FoundSci program under award HR00112490372, DARPA TIAMAT program under award HR00112490419, ARO under award W911NF2410405, ARL STRONG program under awards W911NF2320182, W911NF2220113, and W911NF242021. The authors thank Kuai Yu for dancing with the robot as the human partner in the hardware experiments. The authors also thank Li-Yu Lo, Yuhao Huang, and Yinsen Jia for insightful discussions, and Weizhe Ni, Xinyuan Luo, Jiaxun Liu, Boyuan Wang, and Zijiang Yang for their assistance with the hardware experiments.

\balance

  \bibliographystyle{IEEEtran}

\begin{thebibliography}{10}
\providecommand{\url}[1]{#1}
\csname url@samestyle\endcsname
\providecommand{\newblock}{\relax}
\providecommand{\bibinfo}[2]{#2}
\providecommand{\BIBentrySTDinterwordspacing}{\spaceskip=0pt\relax}
\providecommand{\BIBentryALTinterwordstretchfactor}{4}
\providecommand{\BIBentryALTinterwordspacing}{\spaceskip=\fontdimen2\font plus
\BIBentryALTinterwordstretchfactor\fontdimen3\font minus \fontdimen4\font\relax}
\providecommand{\BIBforeignlanguage}[2]{{\expandafter\ifx\csname l@#1\endcsname\relax
\typeout{** WARNING: IEEEtran.bst: No hyphenation pattern has been}\typeout{** loaded for the language `#1'. Using the pattern for}\typeout{** the default language instead.}\else
\language=\csname l@#1\endcsname
\fi
#2}}
\providecommand{\BIBdecl}{\relax}
\BIBdecl

\bibitem{exbody}
X.~Cheng, Y.~Ji, J.~Chen, R.~Yang, G.~Yang, and X.~Wang, ``Expressive whole-body control for humanoid robots,'' \emph{arXiv preprint arXiv:2402.16796}, 2024.

\bibitem{exbody2}
M.~Ji, X.~Peng, F.~Liu, J.~Li, G.~Yang, X.~Cheng, and X.~Wang, ``{ExBody2}: Advanced expressive humanoid whole-body control,'' \emph{arXiv preprint arXiv:2412.13196}, 2024.

\bibitem{amp}
X.~B. Peng, Z.~Ma, P.~Abbeel, S.~Levine, and A.~Kanazawa, ``{AMP}: Adversarial motion priors for stylized physics-based character control,'' \emph{arXiv preprint arXiv:2104.02180}, 2021.

\bibitem{softmimic}
G.~Margolis, M.~Wang, N.~Fey, and P.~Agrawal, ``{SoftMimic}: Learning compliant whole-body control from examples,'' \emph{arXiv preprint arXiv:2510.17792}, 2025.

\bibitem{kosuge2003msdancer}
K.~Kosuge, T.~Hayashi, Y.~Hirata, and R.~Tobiyama, ``Dance partner robot -- {Ms DanceR},'' in \emph{IEEE/RSJ International Conference on Intelligent Robots and Systems (IROS)}, vol.~3, 2003, pp. 3459--3464.

\bibitem{kobayashi2022boxstep}
T.~Kobayashi, E.~Dean-Leon, J.~R. Guadarrama-Olvera, F.~Bergner, and G.~Cheng, ``Whole-body multicontact haptic human-humanoid interaction based on leader-follower switching: A robot dance of the `box step','' \emph{Advanced Intelligent Systems}, vol.~4, no.~2, p. 2100038, 2022.

\bibitem{charbonneau2024reemc}
M.~Charbonneau, F.~J. {Andrade Chavez}, and K.~Mombaur, ``Dancing with {REEM-C}: A robot-to-human physical-social communication study,'' \emph{arXiv preprint arXiv:2408.05301}, 2024.

\bibitem{dfm}
R.~Watanabe, C.~Li, and M.~Hutter, ``{DFM}: Deep fourier mimic for expressive dance motion learning,'' \emph{arXiv preprint arXiv:2502.10980}, 2025.

\bibitem{rhythm}
H.~Chen, W.~Zhang, P.~Li, S.~Ma, K.~Ma, Y.~Jin, Z.~Xu, X.~Wang, Y.~Zheng, Z.~Wang, J.~Zhao, Y.~Chen, and W.~Ding, ``Rhythm: Learning interactive whole-body control for dual humanoids,'' \emph{arXiv preprint arXiv:2603.02856}, 2026.

\bibitem{cola}
Y.~Du, Y.~Li, B.~Jia, Y.~Lin, P.~Zhou, W.~Liang, Y.~Yang, and S.~Huang, ``Learning human-humanoid coordination for collaborative object carrying,'' \emph{arXiv preprint arXiv:2510.14293}, 2025.

\bibitem{h2compact}
G.~C.~R. Bethala, H.~Huang, N.~Pudasaini, A.~M. Ali, S.~Yuan, C.~Wen, A.~Tzes, and Y.~Fang, ``{H$^2$-COMPACT}: Human-humanoid co-manipulation via adaptive contact trajectory policies,'' \emph{arXiv preprint arXiv:2505.17627}, 2025.

\bibitem{humanx}
Y.~Wang, Q.~Zhao, Y.~F. Lau, R.~Yu, H.~W. Tsui, Q.~Chen, J.~Wang, J.~Pang, and P.~Tan, ``{HumanX}: Toward agile and generalizable humanoid interaction skills from human videos,'' \emph{arXiv preprint arXiv:2602.02473}, 2026.

\bibitem{pair_dstar}
W.-J. Huang, Y.-Y. Zhang, Y.-L. Wei, Z.-W. Xia, J.~Tan, Y.-M. Li, Z.~Zhao, and W.-S. Zheng, ``Learning whole-body human-humanoid interaction from human-human demonstrations,'' \emph{arXiv preprint arXiv:2601.09518}, 2026.

\bibitem{deepmimic}
X.~B. Peng, P.~Abbeel, S.~Levine, and M.~van~de Panne, ``{DeepMimic}: Example-guided deep reinforcement learning of physics-based character skills,'' \emph{arXiv preprint arXiv:1804.02717}, 2018.

\bibitem{ase}
X.~B. Peng, Y.~Guo, L.~Halper, S.~Levine, and S.~Fidler, ``{ASE}: Large-scale reusable adversarial skill embeddings for physically simulated characters,'' \emph{arXiv preprint arXiv:2205.01906}, 2022.

\bibitem{h2o}
T.~He, Z.~Luo, W.~Xiao, C.~Zhang, K.~Kitani, C.~Liu, and G.~Shi, ``Learning human-to-humanoid real-time whole-body teleoperation,'' \emph{arXiv preprint arXiv:2403.04436}, 2024.

\bibitem{omnih2o}
T.~He, Z.~Luo, X.~He, W.~Xiao, C.~Zhang, W.~Zhang, K.~Kitani, C.~Liu, and G.~Shi, ``{OmniH2O}: Universal and dexterous human-to-humanoid whole-body teleoperation and learning,'' \emph{arXiv preprint arXiv:2406.08858}, 2024.

\bibitem{beyondmimic}
Q.~Liao, T.~E. Truong, X.~Huang, Y.~Gao, G.~Tevet, K.~Sreenath, and C.~K. Liu, ``{BeyondMimic}: From motion tracking to versatile humanoid control via guided diffusion,'' \emph{arXiv preprint arXiv:2508.08241}, 2025.

\bibitem{hogan1985impedance}
N.~Hogan, ``Impedance control: An approach to manipulation: Parts {I--III},'' \emph{ASME Journal of Dynamic Systems, Measurement, and Control}, vol. 107, no.~1, pp. 1--24, 1985.

\bibitem{facet}
B.~Xu, H.~Weng, Q.~Lu, Y.~Gao, and H.~Xu, ``{FACET}: Force-adaptive control via impedance reference tracking for legged robots,'' \emph{arXiv preprint arXiv:2505.06883}, 2025.

\bibitem{unifp}
P.~Zhi, P.~Li, J.~Yin, B.~Jia, and S.~Huang, ``Learning a unified policy for position and force control in legged loco-manipulation,'' \emph{arXiv preprint arXiv:2505.20829}, 2025.

\bibitem{ceer}
X.~Luo, X.~Chen, X.~Yin, H.~Wu, B.~Xia, Z.~Chen, J.~Li, B.~Chen, and X.~Cheng, ``{CEER}: Compliant end-effector and root control as a unified interface for hierarchical humanoid loco-manipulation,'' \emph{arXiv preprint arXiv:2605.19981}, 2026.

\bibitem{gentlehumanoid}
Q.~Lu, Y.~Feng, B.~Shi, M.~Piseno, Z.~Bao, and C.~K. Liu, ``{GentleHumanoid}: Learning upper-body compliance for contact-rich human and object interaction,'' \emph{arXiv preprint arXiv:2511.04679}, 2025.

\bibitem{prompthmr}
Y.~Wang, Y.~Sun, P.~Patel, K.~Daniilidis, M.~J. Black, and M.~Kocabas, ``{PromptHMR}: Promptable human mesh recovery,'' \emph{arXiv preprint arXiv:2504.06397}, 2025.

\bibitem{smplx}
G.~Pavlakos, V.~Choutas, N.~Ghorbani, T.~Bolkart, A.~A.~A. Osman, D.~Tzionas, and M.~J. Black, ``Expressive body capture: {3D} hands, face, and body from a single image,'' in \emph{IEEE/CVF Conference on Computer Vision and Pattern Recognition (CVPR)}, 2019, pp. 10\,975--10\,985.

\bibitem{retargeting_matters}
J.~P. Araujo, Y.~Ze, P.~Xu, J.~Wu, and C.~K. Liu, ``Retargeting matters: General motion retargeting for humanoid motion tracking,'' \emph{arXiv preprint arXiv:2510.02252}, 2025.

\bibitem{mjlab}
K.~Zakka, Q.~Liao, B.~Yi, L.~Le~Lay, K.~Sreenath, and P.~Abbeel, ``mjlab: A lightweight framework for {GPU}-accelerated robot learning,'' \emph{arXiv preprint arXiv:2601.22074}, 2026.

\end{thebibliography}

\onecolumn
\section*{Appendix}
\label{sec:appendix}

\Cref{tab:rewards,tab:observations,tab:training_setup} list the reward terms, the observations, and the randomization, force replay and episode settings of the two trained policies.

\begin{table*}[!ht]
\centering
\caption{Reward terms of the two trained policies. Tracking terms are exponential kernels $\exp(-e^2/\sigma^2)$ of the error $e$ with the listed $\sigma$; weights multiply the per-step term. The two policies share every term except the locomotion objective: the decoupled policy draws its locomotion style from the adversarial term, the whole-body tracker (in the style of SoftMimic) from reference tracking of the lower body of the adapted reference.}
\label{tab:rewards}
\footnotesize
\begin{tabular}{@{}lcc>{\raggedright\arraybackslash}p{0.44\textwidth}@{}}
\toprule
Term & Decoupled & Whole-body & Parameters \\
\midrule
\multicolumn{4}{@{}l}{\textit{Partner following}} \\
Right foot follows the partner's left foot & 2 & 2 & offset \SI{0.5}{\meter} behind, $\sigma$ = \SI{0.25}{\meter}, deadband \SI{0.05}{\meter}, height tracked \\
Left foot follows the partner's right foot & 2 & 2 & same \\
Feet air time & 0.5 & 0.5 & air time between \SI{0.2}{\second} and \SI{0.6}{\second} \\
\midrule
\multicolumn{4}{@{}l}{\textit{Upper-body compliance}} \\
Arm positions, anchor-relative & 1 & 1 & $\sigma$ = \SI{0.3}{\meter}; shoulder, elbow, wrist links \\
Arm orientations, anchor-relative & 2 & 2 & $\sigma$ = \SI{0.4}{\radian} \\
Force matching at contacted wrists & 2 & 2 & $\sigma$ = \SI{15}{\newton}, mean over active contacts \\
Torque matching at contacted wrists & 2 & 2 & $\sigma$ = \SI{2}{\newton\meter} \\
Contacted-link position emphasis & 2.5 & 2.5 & $\sigma$ = \SI{0.2}{\meter} \\
Contacted-link orientation emphasis & 2.5 & 2.5 & $\sigma$ = \SI{0.3}{\radian} \\
\midrule
\multicolumn{4}{@{}l}{\textit{Lower body}} \\
Locomotion objective: adversarial term on the lower body & 0.1 (coef.) & & 7 bodies, positions and 6D orientations per frame, 2 frames, pelvis anchor; discriminator 256, 256, gradient penalty 5 \\
Anchor position, global &  & 0.5 & $\sigma$ = \SI{0.3}{\meter} \\
Anchor orientation, global &  & 0.5 & $\sigma$ = \SI{0.4}{\radian} \\
Lower-body positions, anchor-relative &  & 1 & $\sigma$ = \SI{0.3}{\meter}; pelvis, hip, knee, ankle links \\
Lower-body orientations, anchor-relative &  & 1 & $\sigma$ = \SI{0.4}{\radian} \\
Lower-body linear velocities, global &  & 1 & $\sigma$ = \SI{1}{\meter\per\second} \\
Lower-body angular velocities, global &  & 1 & $\sigma$ = \SI{3.14}{\radian\per\second} \\
\midrule
\multicolumn{4}{@{}l}{\textit{Regularization and safety}} \\
Action rate & $-0.1$ & $-0.1$ & squared \\
Joint limit violation & $-10$ & $-10$ & all joints \\
Self collision & $-10$ & $-10$ & contact force above \SI{10}{\newton} \\
Contact with the partner's body & $-25$ & $-25$ & fraction of steps in contact \\
Joint acceleration & $-10^{-5}$ & $-10^{-5}$ & hip and knee joints, squared \\
\bottomrule
\end{tabular}
\end{table*}
\begin{table*}[!ht]
\centering
\caption{Observations of the two trained policies, identical for both. Per-step sizes are stacked over the listed history. Noise is uniform in the listed range and applied to the actor only; the actor's joint positions also carry the per-joint encoder bias of \cref{tab:training_setup}. Both networks are MLPs with hidden layers 512, 256, 128 and ELU activations. The AMP discriminator of the decoupled policy reads 2 frames of 63 features (positions and 6D orientations of 7 lower-body and torso links in the pelvis frame), which the policy itself does not see.}
\label{tab:observations}
\footnotesize
\begin{tabular}{@{}lrrlcc@{}}
\toprule
Term & Per step & History & Actor noise & Actor & Critic \\
\midrule
Base linear velocity (IMU) & 3 & 5 &  &  & \ding{51} \\
Base angular velocity (IMU) & 3 & 5 & $\pm0.2$ & \ding{51} & \ding{51} \\
Joint positions & 29 & 5 & $\pm0.01$ & \ding{51} & \ding{51} \\
Joint velocities & 29 & 5 & $\pm0.5$ & \ding{51} & \ding{51} \\
Previous actions & 29 & 5 &  & \ding{51} & \ding{51} \\
Projected gravity & 3 & 5 & $\pm0.05$ & \ding{51} & \ding{51} \\
Reference anchor orientation, free reference & 6 & 5 & $\pm0.05$ & \ding{51} & \ding{51} \\
Reference anchor position, free reference & 3 & 5 &  &  & \ding{51} \\
Reference anchor pose, served reference & 9 & 5 &  &  & \ding{51} \\
Reference joint positions and velocities, free reference & 58 & 5 &  & \ding{51} & \ding{51} \\
Reference joint positions, served reference & 29 & 5 &  &  & \ding{51} \\
Partner pelvis and knees, robot pelvis frame & 9 & 50 & $\pm0.25$ & \ding{51} & \ding{51} \\
Commanded stiffness per wrist (log), linear and rotational & 4 & 5 &  & \ding{51} & \ding{51} \\
Robot body positions, base frame (14 bodies) & 42 & 5 &  &  & \ding{51} \\
Robot body orientations, base frame (14 bodies) & 84 & 5 &  &  & \ding{51} \\
Applied wrench per wrist & 12 & 5 &  &  & \ding{51} \\
Scheduled wrench per wrist & 12 & 5 &  &  & \ding{51} \\
Environment spring stiffness per wrist (log) & 4 & 5 &  &  & \ding{51} \\
\bottomrule
\end{tabular}
\end{table*}
\begin{table*}[!ht]
\centering
\caption{Randomization, force replay and episode settings, identical for both policies.}
\label{tab:training_setup}
\footnotesize
\begin{tabular}{@{}>{\raggedright\arraybackslash}p{0.34\linewidth}>{\raggedright\arraybackslash}p{0.62\linewidth}@{}}
\toprule
Setting & Value \\
\midrule
\multicolumn{2}{@{}l}{\textit{Randomization (at episode start)}} \\
Torso centre of mass offset & uniform, $x$ $\pm$\SI{0.025}{\meter}, $y$ $\pm$\SI{0.05}{\meter}, $z$ $\pm$\SI{0.05}{\meter} \\
Foot friction coefficient & uniform, \numrange{0.3}{1.2}, one draw for all foot geoms \\
Joint encoder bias & uniform, $\pm$\SI{0.01}{\radian} per joint, added to the actor's joint positions \\
Actor observation noise & uniform, per observation term, at the ranges the policy trained with \\
Reset root pose & $x$, $y$ $\pm$\SI{0.15}{\meter}, yaw $\pm$\SI{0.3}{\radian} about the clip pose \\
\midrule
\multicolumn{2}{@{}l}{\textit{Force replay}} \\
Environment spring & stiffness and setpoint from the force schedule, force scale 1 \\
Setpoint frame & torso frame frozen at the rising edge of each event, yaw only \\
Pushes & none besides the scheduled events \\
\midrule
\multicolumn{2}{@{}l}{\textit{Episode}} \\
Length, control rate & \SI{20}{\second}, \SI{50}{\hertz} \\
Clip end & clip-chaining to another clip of the training data \\
Termination & anchor height below \SI{0.3}{\meter}, anchor orientation off by more than \SI{1.2}{\radian}, a foot more than \SI{1.2}{\meter} from its target, time out \\
Parallel environments & 16384 \\
\bottomrule
\end{tabular}
\end{table*}

\end{document}